\documentclass{article}
\usepackage{spconf,amsmath,graphicx}
\usepackage{booktabs}
\usepackage{float}
\usepackage{verbatimbox}
\usepackage{amsfonts,amssymb}
\usepackage{algorithmic}
\usepackage{textcomp}
\usepackage{array} 
\usepackage{subfigure}
\usepackage{setspace}
\usepackage{amssymb}
\usepackage{booktabs}
\usepackage{stfloats}
\usepackage{autobreak}
\usepackage{threeparttable}
\usepackage{wrapfig}
\usepackage{multirow}
\usepackage{hhline}
\usepackage{bbm}
\usepackage[misc]{ifsym}

\title{Consistent Posterior Distributions Under Vessel-Mixing: A Regularization for Cross-Domain Retinal Artery/Vein Classification}

\name{Chenxin Li, Yunlong Zhang, Zhehan Liang, Wenao Ma, \Letter Yue Huang, Xinghao Ding
\thanks{The work is supported in part by National Natural Science Foundation of China under Grants 81671766, U19B2031, in part by Open Fund of Science and Technology on Automatic Target Recognition Laboratory 6142503190202, in part by Fundamental Research Funds for the Central Universities 20720200003, in part by the Science and Technology Key Project of Fujian Province, China (No. 2019HZ020009).
Corresponding author: Yue Huang, yhuang2010@xmu.edu.cn}
} 
\address{
\textsuperscript{1}Key Laboratory of Underwater Acoustic Communication and Marine Information Technology, \\
Minister of Education, Xiamen University, Xiamen, China\\
\textsuperscript{2}School of Informatics, Xiamen University, Xiamen, China
}

\begin{document}
\maketitle   
\begin{abstract}
Retinal artery/vein (A/V) classification is a critical technique for diagnosing diabetes and cardiovascular diseases. Although deep learning based methods achieve impressive results in A/V classification, the performance usually degrades when directly apply the models that trained on one dataset to another set, due to the domain shift, e.g., caused by the variations in imaging protocols.
In this paper, we propose a novel method to improve cross-domain generalization for pixel-wise retinal A/V classification. That is, vessel-mixing based consistency regularization, which regularizes the models to give consistent posterior distributions for vessel-mixing samples.
The proposed method achieves the state-of-the-art performance on extensive experiments for cross-domain A/V classification, which is even close to the performance of fully supervised learning on target domain in some cases.
\end{abstract}   
\section{Introduction}
\label{sec:Introduction}
The clinical features in retinal arteries and veins (A/V) serve as biomarkers in diagnosing many systemic and cardiovascular diseases. 
Therefore, accurate automatic pixel-wise classification of retinal A/V in fundus images is highly desired.
Conventional machine learning based methods usually start from a pre-segmented vascular tree, and then extract hand-crafted features or incorporate the graphical connection information to learn a classifier for vessel pixels \cite{xu2017improved,zhao2019retinal}. Recently, several deep learning based works simultaneously segment and classify retinal A/V \cite{meyer2018deep,ma2019multi}, employing the deep semantic-segmentation models that are trained on the well-annotated fundus images.   

\begin{figure}[!h]
\centering        
\includegraphics[width=0.8\linewidth,height=2.3cm]{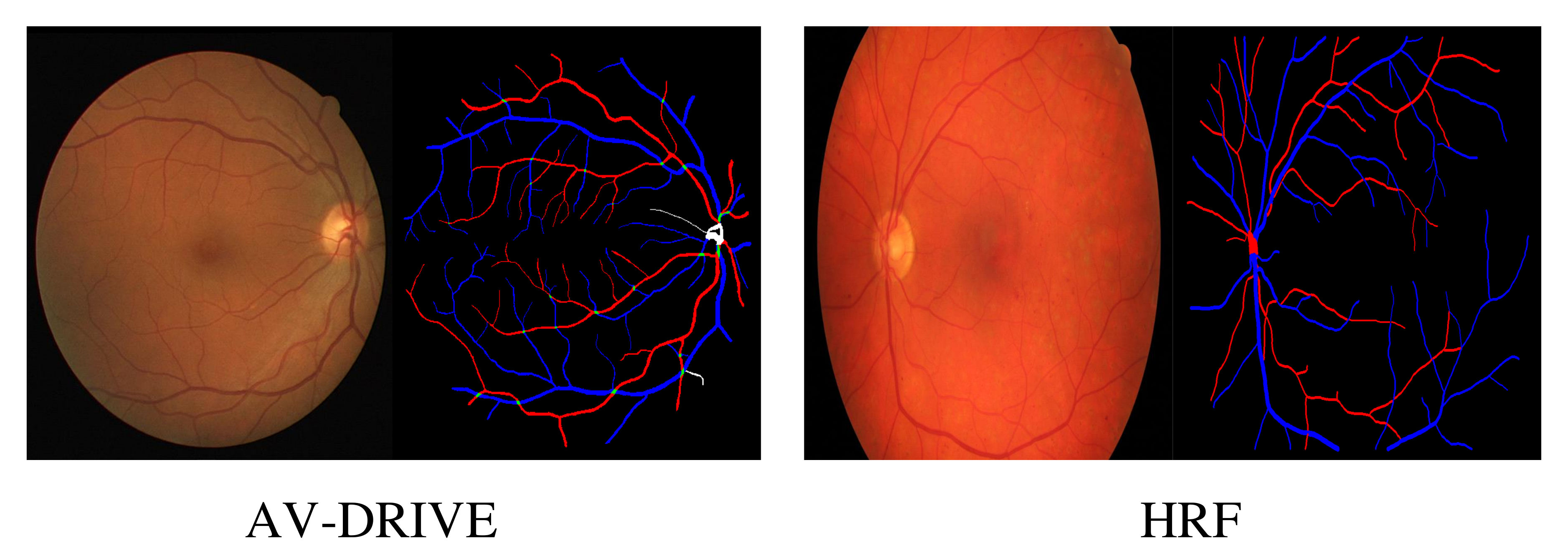}
\begin{small}
\vspace{-0.25cm}
\caption{Retinal fundus images and A/V annotations from diverse sets. Color R/B/G denotes artery/vein/crossing labels.} \label{fig_dataset}
\end{small}
\end{figure}

Despite their success, most of them neglect the potential domain shift between training and testing dataset.
An example is using the model trained on the fundus images from one medical institution, to segment A/V on the images from another institution. The datasets are collected from
different institutions and imaging devices, and appear in various field-of-view, resolution and light intensity \cite{galdran2019uncertainty} (cf. Fig.\ref{fig_dataset}). There exists a non-negligible domain shift across the datasets, which causes the cross-domain data and the degradation of evaluation performance.  
To overcome the challenge, a straightforward technique is fine-tuning the trained model with additional annotated data from target domain. 
But the cost of manually labeling for each target domain is expensive, especially for the retinal vessels with complicated structures. 

Without the extra need for target-domain labels,
recent unsupervised domain adaptation (UDA) seems appealing in medical image segmentation \cite{kamnitsas2017unsupervised,dou2018pnp, perone2019unsupervised,liu2019cfea}.    
Among them, the current mainstream extracts domain-invariant features by minimizing the inter-domain discrepancy, through the distance loss of feature, or adversarial discriminative learning. However, as discussed in \cite{zhang2017curriculum,liu2019cfea}, different from classification task,
{\itshape forcing the alignment of high-dimensional feature maps in segmentation task may also harm the contained structure information, and further causes negative transfer} (c.f. Tab. \ref{tab1}). This could be worse for retinal A/V in fundus images where the complicated graphical structures of vessels contain more refined structure information (c.f. Fig. \ref{fig_dataset}).
\vspace{-0.2cm}

\vspace{-0.05cm}
In this paper, we propose a novel vessel-mixing based consistency regularization framework to emphasize generalization under domain shift, for cross-domain retinal A/V classification.
The proposed method regularizes models to give consistent posterior distributions of A/V for the inputs under vessel-mixing perturbation, which is generated by regionally mixing the vessel structures from two fundus images. 
Extensive experiments on cross-domain A/V classification is conducted, using four public fundus images sets from different institutions. 
The proposed method achieves promising results, e.g., in the HRF$\to$DRIVE task, our method recovers the degraded F1 score from 79.5\% to 87.6\%, which is also very close to the 89.1\% F1 achieved by supervised learning.

\begin{figure*}[!h]     
	\centering
	\includegraphics[width=0.95\linewidth,height=7.8cm]{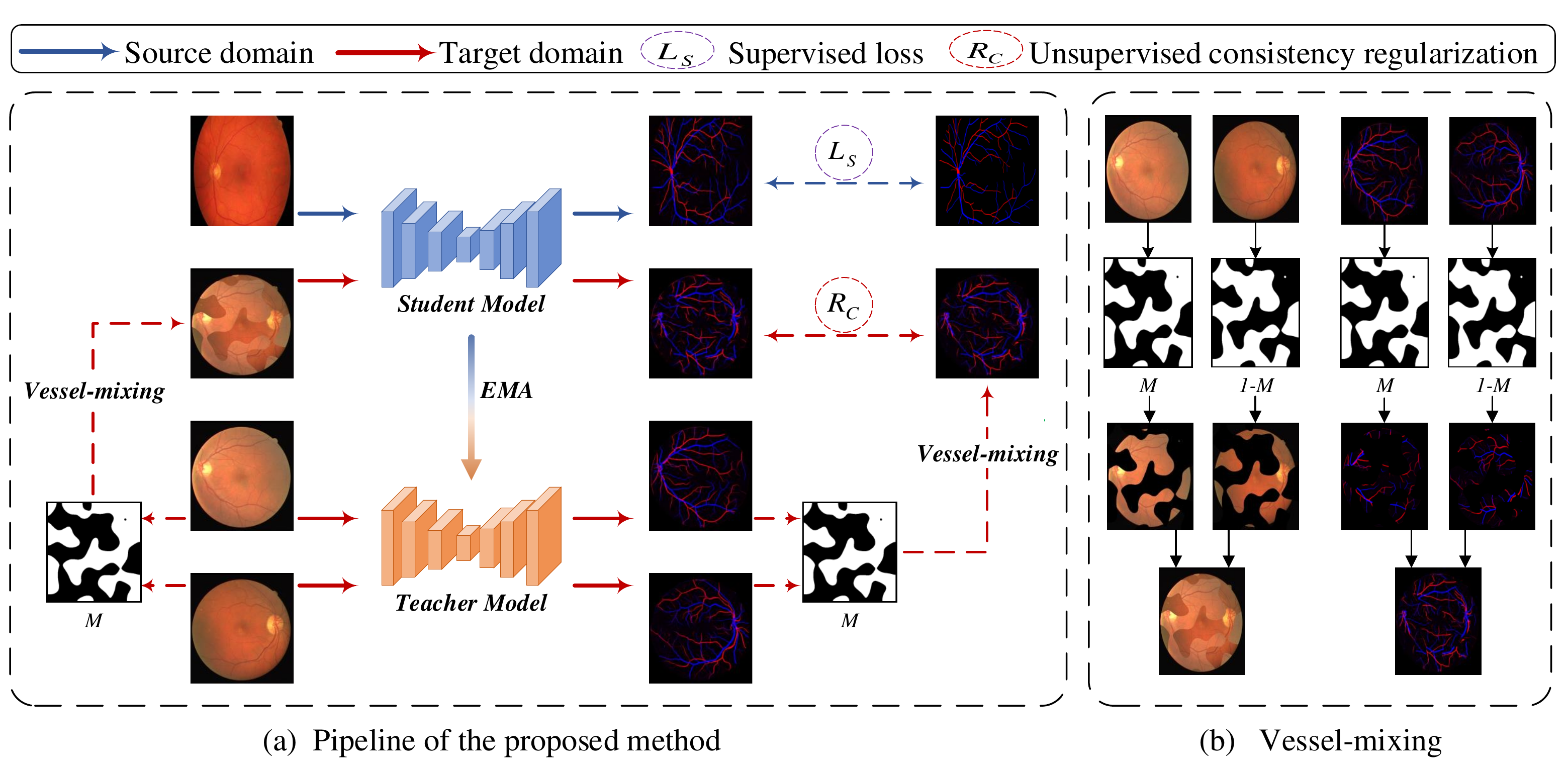}  
	\caption{The pipeline of the proposed vessel-mixing based consistency regularization. {\itshape EMA} means the exponential moving average of the student model's weights are transferred to the teacher model's weights \cite{tarvainen2017mean}. {\it Vessel-mixing} is described in (b).} \label{fig_pipline}
\end{figure*}


\section{Method}
\label{sec:Method}
\subsection{Overview}
First of all, we assume that we are given a set of fundus images $x_s$ along with their pixel-wise A/V annotations $y_s$, and a set of unannotated fundus images $x_t$. The former is regarded as the source domain and the latter as the target domain.
As shown in Fig. \ref{fig_pipline}, the training pipeline of proposed framework is based on a supervised segmentation loss on the labeled source domain, and an unsupervised regularization term on the unlabeled target domain, i.e., the proposed vessel-mixing based consistency regularization. This helps models withstand the domain shift, and improves cross-domain generalization.
At a high level, the proposed regularization is characterized by its utilization of vessel-mixing perturbations in concert with a consistency loss. The vessel-mixing operations are generated from regionally mixing the vessel structures from two fundus images.
We then enforce the consistent pixel-level predictions by models between the origin input images and vessel-mixing perturbated ones, through the use of mean-squared error (MSE) distance as a consistency loss.   

One realization of the consistency regularization is following $\Pi$-model \cite{laine2016temporal}, which passes each unlabeled sample through a model twice. Another is the mean teacher framework \cite{tarvainen2017mean}, which encourages consistency between predictions of a student model and a teacher model whose weights are an exponential moving average (EMA) of those of the student. 
In this paper, we utilize the mean teacher framework, which is reported to perform better \cite{tarvainen2017mean}, as the backbone following the previous works \cite{perone2019unsupervised,liu2019cfea}. To distinguish, we denote the student model as $f_{\theta}$ and the teacher model as $f_{\theta^{\prime}}$.

\subsection{Vessel-mixing perturbation}
The data augmentation or perturbation is an essential component in the strategy of consistency regularization for inducing robustness \cite{tarvainen2017mean,laine2016temporal,2019arXiv190601916F}. Previously, the related works \cite{perone2019unsupervised,liu2019cfea} in the UDA community of medical segmentation employ spatial transformation based perturbations such as flipping, scaling, rotation, etc. Recently, the mixup \cite{zhang2017mixup} is proposed as a powerful technique of augmentation for the classification task by mixing two images globally. But how to adapt it in the pixel-wise classification task remains a problem.
In our method, we expand the mix-based perturbation from the task of image-wise classification to pixel-wise classification by extending mixing from a global level to a regional level.
We propose vessel-mixing, a task-driven perturbation, to generate the diverse mixtures of retinal vessels from two fundus images.

Concretely, to deploy the vessel-mixing, we first generate a guidance mask $M$ in the following ways. We sample a Gaussian random matrix with the same size as the images to be mixed, apply a Gaussian filter to introduce regional correction, and then binarize the pixel values with a threshold, which is the mean value of the matrix in our case. 
As shown in Fig. \ref{fig_pipline}, given the produced binary mask $M$ and the input target-domain image $x_{t1}$ and $x_{t2}$, the result vessel-mixing image is obtained, formulated as:
\begin{small}
\begin{equation}
\label{eq1}
	Mix(x_{t1},x_{t2},M)=M\odot x_{t1}+(\mathbbm{1}-M)\odot x_{t2}
\end{equation}
\end{small}
where $\odot$ represents Hadamard product. Then the result is then fed into the student model as shown in Fig. \ref{fig_pipline} (a). Present in Fig. \ref{fig_pipline} (b), the proposed vessel-mixing generates the result as the regional mixtures of two input fundus images and perturb the structures of retinal vessels to a certain extent. Moreover, with the mask $M$ randomly sampled continuously, the vessel-mixing allows us to produce diverse perturbation in fundus images, especially the retinal vessels. Similarly, we also carry out the vessel-mixing in the output of the teacher model, i.e., the segmentation maps of input image $x_{t1}$ and $x_{t2}$ respectively, formulated as:
\begin{small}
\begin{equation}
	\label{eq2}
	Mix(f_{\theta^\prime}(x_{t1}),f_{\theta^\prime}(x_{t2}),M)=M\odot f_{\theta^\prime}(x_{t1}) + (\mathbbm{1}-M)\odot f_{\theta^\prime}(x_{t2})
\end{equation}
\end{small}
\vspace{-0.5cm}
\subsection{Consistency regularization}
We couple with this vessel-mixing perturbation scheme a loss that enforces smoother neural network responses. 
Applied with the vessel-mixing, although the structure information of retinal A/V is affected, the pixel-level information is approximately preserved. Thus, we would like the model to embed the origin input image $x_{1t}$ and $x_{2t}$ and the mixture $Mix(x_{t1},x_{t2},M)$ similarly pixel by pixel. Toward this end, we minimize the mean-squared error (MSE) loss among the the mixture of the posterior distributions of the origin input images $x_{1t}$ and $x_{2t}$, and the posterior distributions of their mixed variant $Mix(x_{t1},x_{t2},M)$, formulated as:
\begin{small}
\begin{equation}
\label{eq3}
\mathcal{R}_C=\mathcal{L}_{mse}(f_\theta(Mix(x_{t1},x_{t2},M)),Mix(f_{\theta^\prime}(x_{t1}),f_{\theta^\prime}(x_{t2}),M))
\end{equation}
\end{small}
Then, the origin supervised segmentation loss $\mathcal{L}_S$ is replaced with a combined term, formulated as:
\begin{small}
\begin{equation}
\label{eq4}
\mathcal{L}=\mathcal{L}_S+\lambda \mathcal{R}_C
\end{equation}
\end{small}
where $\lambda$ is the trade-off parameter that controls the training bias to the regularization term. We hope it can be self-adaptive according to the progress of training. When the models assign the target-domain data with higher confidence, we think the consistency regularization can be more reliable. So we update $\lambda$ by the averaged threshold confidence, which is achieved by calculating the proportion of pixel-level predictions over the threshold in the prediction maps of target-domain data (i.e., the output of teacher model).

Additionally, for the supervised segmentation loss in this paper, we employ the sum of cross-entropy (BCE) losses, respectively for arteries and veins,
formulated as:
\begin{small}
\begin{equation}
\label{eq5}
\mathcal{L}_S=\mathcal{L}_{bce}^{a}(f_\theta(x_s),y_s,w_{pos})+\mathcal{L}_{bce}^{v}(f_\theta(x_s),y_s,w_{pos})
\end{equation}
\end{small}
where $a$ and $v$ denotes A/V. $w_{pos}$ is a positive weight set as 10 to alleviate the class imbalance.
The overall training procedure is as following. In each iteration, the student model is optimized by minimizing the combined total loss $\mathcal{L}$ described in Eq. \ref{eq4}. And the parameters of the teacher model are updated as the exponential moving average (EMA) of those the student model, which is described above and formulated as:
\begin{small}
\begin{equation} 
\label{eq6}
\theta_{t}^{\prime}=\alpha\theta_{t-1}^{\prime}+(1-\alpha)\theta_{t}
\end{equation}
\end{small}
where $t$ and $t-1$ represent the training iteration, and $\alpha$ denotes the smoothing coefficient hyperparameter in EMA, which is 0.99 according to the common setting \cite{tarvainen2017mean,perone2019unsupervised,liu2019cfea}.

\section{Experiments}
\subsection{Dataset}
We evaluate the proposed method by using four public color retinal fundus image datasets. These include AV-DRIVE  \cite{DRIVE}, High Resolution Fundus (HRF) \cite{HRF}, LES-AV \cite{LES} and INSPIRE-AVR \cite{INSPIRE}. These datasets are collected by different institutions and devices, causing the domain shift among the datasets. The AV-DRIVE dataset has 20/20 (training/testing) images with the resolution of 584$\times$565, and publicly available pixel-wise labeling of A/V classification. The HRF database contains 45 images of dimensions 1200$\times$800, with a regular split of 23/22 and pixel-level annotations. The LES-AV dataset contains 22 optic-disc centered images with a resolution of 1620$\times$1444 pixels, with a split of 11/11.
INSPIRE-AVR contains 40 images of size 2048$\times$2392, with A/V labels available only for centerline rather than pixel-wise annotations, so it can only be used for evaluation. For simplicity, we respectively denote: AV-DRIVE (D), HRF (H), LES-AV (L), INSPIRE-AVR (I).

\subsection{Implementation details}
We employ the U-Net architecture as the backbone, for both the teacher model and student model.
In the training stage, we randomly sample a new mask $M$ in every iteration.
we adopt Adam optimizer with a constant learning rate of 0.001 to optimize the student model. Mini-batch size is 2 for both source and target domains.
We resize the input images to a common resolution of 512$\times$512, and normalize them to $[0,1]$
The standard data augmentation is applied to avoid overfitting, including random flipping, rotating, and scaling.
The total training iterations is 10000.
In the testing stage, we use only the teacher model without $Mix$ module. We implement the experiments in Pytorch, an NVIDIA Titan Xp GPU.

\begin{figure}[htbp]
	\centering
	\includegraphics[width=0.92\linewidth,height=4.6cm]{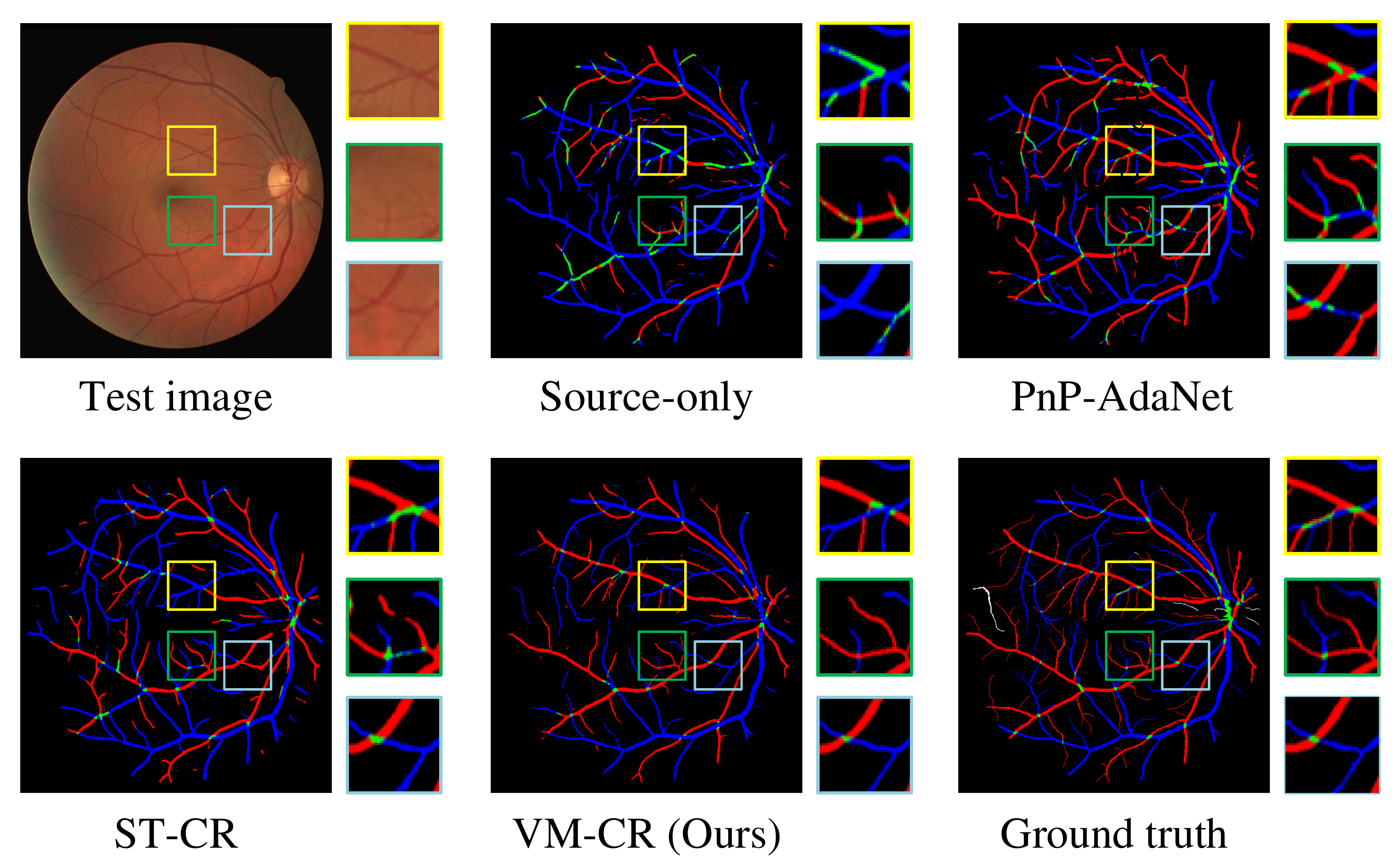}
	\vspace{-0.2cm}
	\caption{Visual results of the UDA methods on cross-dataset A/V classification for the H$\to$D task.} 
	\label{fig_H2D} 
\end{figure}
\vspace{-0.2cm}

\subsection{Experimental results}
We adopt average accuracy (Acc), sensitivity (Sen), specificity (Sp) and F1-score (F1) for quantitative evaluation.
Besides, following \cite{ma2019multi}, we evaluate A/V classification
on the ground truth vessel pixels rather than the segmented vessels, which is more strict since some capillary vessels may not be well segmented.
We report the results of the proposed method on four cross-database experiments in Tab. \ref{tab1}. Also, we compare with several UDA-segmentation methods, AdvNet \cite{kamnitsas2017unsupervised}, PnP-AdaNet \cite{dou2018pnp} and ST-CR \cite{perone2019unsupervised}. We use the publicized codes provided by their authors. 
Among them, the first two are the recently proposed domain-align based methods. ST-CR and VM-CR stand for the spatial transformation based consistency regularization and the proposed vessel-mixing consistency regularization, respectively, and the main difference is the utilized perturbation. Besides, UA-AV \cite{galdran2019uncertainty} is a recent work specifically for cross-domain retinal A/V classification and trained on only AV-DRIVE.

Firstly, we observe that {\itshape Source-only} suffers heavy performance degradation compared to {\itshape Target-only}. For example, in L$\to$D, the degradation performance gap is 13.9\% at F1, 15.0\% at Acc, 22.0\% at Sp, showing the existence of domain shifts.
The domain alignment methods achieve some improvements in several tasks, but AdaNet achieves even worse F1 score than {\itshape Source-only} in L$\to$D, confirms our concern about {\itshape negative transfer that may be caused by enforcing the high-dimensional feature maps of semantic segmentation task}. Instead, the consistency regularization based methods achieve significant improvements. Moreover, the proposed VM-CR achieves the state-of-the-art performance in all the four cross-database tasks, and even performs very close to {\itshape Target-only}.
For example, we improve the F1 score to 87.6\% and Acc to 89.1\%, in H$\to$D task, where {\itshape Target-only} achieves 89.1\% for F1 and 90.2\% for Acc. 
Fig. \ref{fig_H2D} shows the visual results in H$\to$D task. The proposed method segments and classifies most of the vessels accurately, which significantly improves the classification of A/V over {\itshape Source-only}, especially for the marked challenging regions.

\begin{figure}[!h]
	\centering
	\includegraphics[width=0.95\linewidth,height=2.3cm]{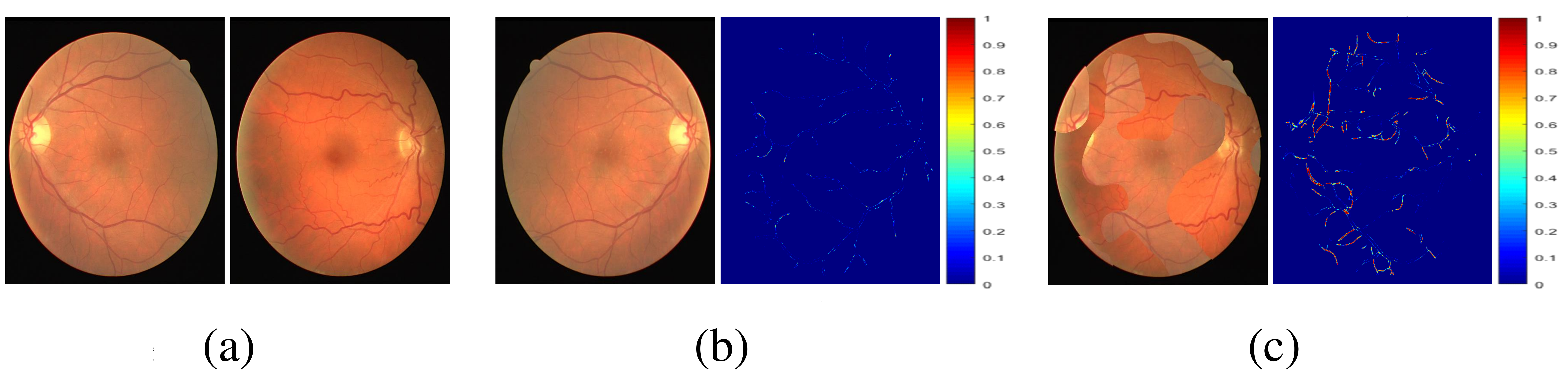}
	\vspace{-0.3cm}   
	\caption{The visualization of consistency loss maps (pixel-level squared-error losses between predictions for perturbated input and perturbated predictions for origin input) 
	(a): input image A and B. (b): input A under spatial transformation, i.e., flip, and corresponding consistency loss map. (c): input A and B under the proposed vessel-mixing, and consistency loss map.
} \label{fig_lossmapshow}      
\end{figure}

As shown in Fig. \ref{fig_lossmapshow}, to further highlight the effectiveness of the proposed vessel-mixing component, we observe the consistency loss maps in the early training stage. 
The numeric values of the loss map from vessel-mixing is significant higher than spatial transformation, especially in the vessel structures, which means the vessel-mixing is a stronger perturbation and thus pushes the model to be more robust in turn under the consistency regularization.

\begin{table}[!t]\small
	\centering
	\caption{The results of A/V classification on four cross-database tasks. $\to$ denotes {\itshape Source} adapted to {\itshape Target}. {\itshape Source-only}/{\itshape Target-only} denotes the evaluation results of models supervised trained on source/target data. 
	{\itshape Target-only} in D$\to$I is lacking since INSPIRE has only centerline annotations that can't be used for supervised learning. 
	Please note that here we evaluate A/V-classification performance on the ground truth vessels pixels, which is a more strict criterion, following \cite{ma2019multi,galdran2019uncertainty}. 
	$\star$ denotes the observed negative transfer.
	} \label{tab1}
	\begin{tabular}{p{0.26\linewidth}|p{0.31\linewidth}<{\centering}|p{0.31\linewidth}<{\centering}}
		\hline
		\multirow{2}*{Method}&\multicolumn{1}{c|}{H$\to$D}&\multicolumn{1}{c}{L$\to$D} \\
		\cline{2-3}
		&F1/Acc/Sen/Sp & F1/Acc/Sen/Sp   \\
		\hline
		Source-only&79.5/83.0/76.2/88.6  &75.2/75.2/82.4/69.2   \\
		\hline
		AdvNet \cite{kamnitsas2017unsupervised}&81.0/83.2/81.0/85.3 &$\star$74.9/76.5/77.8/75.4 \\
		PnP-AdaNet\cite{dou2018pnp}&84.3/85.8/85.0/86.6
		&77.7/79.7/78.4/80.9    \\
		ST-CR \cite{perone2019unsupervised}&86.2/87.8/86.6/88.9	&83.0/84.1/\textbf{85.7}/83.1  \\
		\textbf{VM-CR(Ours)}&\textbf{87.6}/\textbf{89.1}/\textbf{86.8}/\textbf{91.1}  &\textbf{84.3}/\textbf{85.5}/85.6/\textbf{85.7}  \\
		\hline
		Target-only&89.1/90.2/89.1/91.2& 89.1/90.2/89.1/91.2  \\
		\hline
	\end{tabular}
	\\[8pt]
	\begin{tabular}{p{0.26\linewidth}|p{0.31\linewidth}<{\centering}|p{0.31\linewidth}<{\centering}}
		\hline
		\multirow{2}*{Method}&\multicolumn{1}{c|}{D$\to$L}&\multicolumn{1}{c}{D$\to$I} \\
		\cline{2-3}
		&F1/Acc/Sen/Sp & F1/Acc/Sen/Sp   \\
		\hline 
		Source-only &79.5/82.0/77.5/85.6  &80.5/80.0/82.2/78.6 \\
		\hline
		UA-AV \cite{galdran2019uncertainty} &86.0/86.0/88.0/85.0  &80.0/80.0/82.0/87.0 \\
		AdvNet \cite{kamnitsas2017unsupervised} &80.6/81.6/83.8/80.0    &84.8/84.1/83.4/85.5 \\
		PnP-AdaNet\cite{dou2018pnp}&83.2/85.3/80.4/89.6  &86.9/86.0/84.6/88.0  \\
		ST-CR \cite{perone2019unsupervised}&86.1/86.8/\textbf{89.7}/84.6 &91.3/90.6/88.4/93.0\\
		\textbf{VM-CR(Ours)}  &\textbf{87.5}/\textbf{88.5}/88.2/\textbf{88.9}  &\textbf{92.5}/\textbf{92.2}/\textbf{94.5}/\textbf{90.5} \\
		\hline
		Target-only  &89.1/89.5/90.3/88.9   &~---~/~---~/~---~/~---~ \\
		\hline
	\end{tabular}

\end{table}
\vspace{-0.15cm}

\vspace{-0.15cm}
\vspace{-0.1cm}

\vspace{-0.25cm}
\section{Conclusion}
In this paper, we propose a novel vessel-mixing based consistency regularization framework, for cross-domain retinal A/V classification. 
The consistency regularization framework constrains the model to give consistent A/V segmentation maps for the unlabeled target inputs, under the vessel-mixing perturbation that regionally mixes two fundus images.
The proposed approach achieves state-of-the-art performance on the extensive experiments of cross-database A/V classification. 

\bibliographystyle{IEEEbib}
\bibliography{ref_V3}
\end{document}